\documentclass{icslp}
\usepackage{leqno}
\usepackage{psfig}

\newenvironment{example}{\begin{examples}\item}{\end{examples}}
\newcounter{equationsave}          
\newenvironment{examples}
{%
\begin{list}{(\theequation)}%
{%
\setcounter{equationsave}{\arabic{equation}}%
\usecounter{equation}
\setcounter{equation}{\arabic{equationsave}}
\setlength{\listparindent}{0pt}%
\def\makelabel##1{##1\hfil}
}%
\raggedright}
{\end{list}}

\title{%
  TOWARDS UNDERSTANDING SPONTANEOUS SPEECH:\\
  WORD ACCURACY VS. CONCEPT ACCURACY
}
\author{%
M.~Boros$^1$, W.~Eckert$^2$, F.~Gallwitz$^2$,\\
G.~G{\"o}rz$^1$, G.~Hanrieder$^1$, H.~Niemann$^{1,2}$
}
\begin{document} 
\affiliation {$^1$Bayerisches Forschungszentrum f\"ur Wissensbasierte
Systeme (FORWISS),\\ Am Weichselgarten 7, D-91058 Erlangen, Germany\\
$^2$Lehrstuhl f\"ur Mustererkennung (Informatik 5),Universit\"at
Erlangen--N\"urnberg,\\ Martensstra{\ss}e 3, D-91058 Erlangen,
Germany\\
email: {\tt boros@forwiss.uni-erlangen.de}} 

\maketitle

\begin{abstract}
In this paper we describe an approach to automatic evaluation of both
the speech {\em recognition} and {\em understanding} capabilities of a
spoken dialogue system for train time table information.  We use {\em
word accuracy} for recognition and {\em concept accuracy} for
understanding performance judgement.  Both measures are calculated by
comparing these modules' output with a correct reference answer.  We
report evaluation results for a spontaneous speech corpus with about
10000 utterances.  We observed a nearly linear relationship between
word accuracy and concept accuracy.
\end{abstract}

\section{INTRODUCTION}
Total system evaluation plays an important role for developers of
spoken dialogue systems, because it allows both to monitor progress
within a single project and to compare different solutions for the
same task.  An objective and verifiable judgement of system
performance requires that the scientific community agrees upon widely
accepted evaluation measures.  In speech recognition, such a mutually
agreed upon measure is available with the so-called {\em word
accuracy} (WA).  There exist standardized tools which can
automatically compute the WA of recognition results for a given test
corpus annotated with transcriptions of the actually spoken
words. This high standard of automatic evaluation methods could not
yet be transferred to the higher processing level of speech {\em
understanding}, although the last few years have witnessed increasing
efforts in the development of an evaluation methodology for natural
language processing (cf.~\cite{HirschmanThompson:96},
\cite{EAGLESEVAL:94},
\cite{Crouch:96}).

This paper describes our approach to automatic evaluation of both the
{\em recognition} and the {\em understanding} capabilities of a spoken
dialogue system for train time table inquiries~\cite{Ecketal:93}. Such
an integrated evaluation environment allows a systematic investigation
of the relationship between recognition and understanding performance.
The central question is: {\em How does a change in the recognition
accuracy affect the understanding accuracy?}  First we describe the
evaluation measures {\em word accuracy} and {\em concept accuracy}.
After this we show our evaluation architecture for automatic
calculation of recognition and understanding accuracy.  Finally, we
report results for a spontaneous speech corpus containing about 10000
utterances.

\section{EVALUATION MEASURES}
\label{EvalMeas}
Automatic evaluation methods require the use of prepared test corpora
in which each test case is combined with a ``correct'' reference
answer against which the system output can be judged.  In speech
recognition, it is relatively uncontroversial how this reference
answers look like: they are transcriptions of the words that were
actually spoken.\footnote{There are still debates on the transcription
and evaluation of spontaneous speech containing fragmentary words,
hesitations, background noise, etc.}  It is less clear, however, what
constitutes the ``correct'' analysis at the level of language
understanding. Currently, there is no agreement among computational
linguists regarding a ``correct'' semantic representation for a wide
variety of linguistic phenomena.  As a consequence, there are no
semantically annotated corpora available as a common test bed for
comparative evaluation of linguistic processing components.
Nevertheless, we believe that an objective and verifiable measurement
of the understanding capabilities of a system can only be achieved
with a ``reference answer''-based approach using test corpora with
semantic annotations.  This conviction is based on the fact that the
main task of the linguistic processing component in a spoken dialogue
system is to map the spoken input to a semantic representation.
Evaluation approaches which look only at the surface
forms\footnote{In~\cite{Schmid:94} a word graph parser is rated by
calculating the {\em sentence recognition accuracy}, which is defined
as ``the number of word graphs where the analysis found the spoken
sentences divided by the number of word graphs''.} or the syntactic
structures~\cite{Black:91} of the parsing results cannot judge the
parser performance regarding the construction of a semantic
representation.  Therefore, we defined a semantic annotation format
within our task domain.  For measuring the understanding performance
we adopted the so-called {\em concept accuracy}.  This measure, which
was proposed from the evaluation working group of the ESPRIT project
SUNDIAL~\cite{SimpsonFraser:93}, can be calculated automatically in
analogy with the recognition measure {\em word accuracy}.

\subsection{Word Accuracy}
Word Accuracy (WA) is a widely accepted evaluation measure for word
recognizers. The automatic calculation of WA for a given set of
recognition results requires the existence of reference
transliterations for all spoken utterances. The reference answers
consist of a transcription of what was actually spoken. Given the
reference REF, the WA of the recognizer output HYP is determined by
calculating the Levenshtein distance 
between REF and HYP and by assigning equal costs to substitution,
insertion, and deletion errors. WA is calculated as a percentage using the
formula 
\begin{equation}
\label{Wortakkuratheit}
 WA = 100\left(1-\frac{W_S+W_I+W_D}{W} \right ) \%
\end{equation}
where $W$ is the total number of words in REF, and $W_S$, $W_I$, $W_D$
are the number of reference words which were substituted, inserted,
and deleted in HYP, respectively.

For example, the WA of the recognized string in~(\ref{WAEx}) is
66.7\%, since the spoken word {\sf I} was deleted and the spoken word
{\sf Berlin} was substituted by {\sf Bonn} in HYP, such that $W_D= 1$
and $W_S = 1$.  By inserting these values into
formula~(\ref{Wortakkuratheit}) the WA is calculated by
$100\left(1-\frac{2}{6} \right )=66.7 \%$.
\begin{example}
\label{WAEx}
\begin{tabular}{|l|rll|}\hline
REF: & {\sf I} & {\sf want to go to} & {\sf Berlin}\\
HYP: & & {\sf want to go to} & {\sf Bonn} \\ \hline
\end{tabular}
\end{example}

\subsection{Concept Accuracy}
\label{SSSA}
While WA evaluates the performance of the speech {\em recognition}
component, the language {\em understanding} capabilities of a system
can be judged by concept accuracy
(CA).%
\footnote{In~\cite{SimpsonFraser:93} a similar measure was called {\em
information content}.}  This approach is based on the assumption that
the main task of the linguistic processor in a spoken dialogue system
is to extract the propositional content from the spoken
utterance. Furthermore, it is assumed that this propositional content
can be represented as a list of {\em semantic units} ({\em SU)} taking
the form of attribute-value pairs.  The definition of the attributes
relevant for understanding is determined by domain-dependent {\em task
parameters} which reflect the functionality of the system.  For
example, in a train time table information task the system cannot
access the connected database system without knowing the values for
the task parameters {\tt sourcecity}, {\tt goalcity} and {\tt date}.
Accordingly, the propositional content of a sentence
like~(\ref{Ex1Satz}) is represented as the series of SUs shown
in~(\ref{Ex1Sem}).
\begin{example}
\label{Ex1Satz}
{\sf I want to go from Bonn to Berlin.}
\end{example}\vspace*{-2em}
\begin{example}
\label{Ex1Sem}
{\tt [sourcecity:Bonn, goalcity:Berlin]}
\end{example}
Given such semantic reference answers in form of task parameter-value
pairs the performance of a speech understanding component can be
measured in analogy with the method used for word recognition
evaluation. Concept accuracy CA can be calculated by replacing the
words $W$ in formula~(\ref{Wortakkuratheit}) with semantic units $SU$:
\begin{equation}
CA = 100\left(1-\frac{SU_S+SU_I+SU_D}{SU}\right) \%
\end{equation}
$SU$ is the total number of semantic units in the reference answer and
$SU_S$, $SU_I$, and $SU_D$ are the number of semantic units that were
substituted, inserted, and deleted in the parser output, respectively.
The calculation of CA will be illustrated in the following example:
\begin{example}
\label{ExSA}
\begin{tabular}{|l|ll|}
\hline
Spoken: & {\sf No} & {\sf to Bonn}\\
REF:    & {\tt dm\_marker:no} & {\tt goalcity:Bonn}\\
\hline
Recog.: &  {\sf No} & {\sf to Berlin} \\
HYP:  & {\tt dm\_marker:no} & {\tt goalcity:Berlin}\\
\hline
\end{tabular}
\end{example}
The total number of uttered semantic units in~(\ref{ExSA}) is
$SU=2$. Due to the misrecognition of the spoken word {\sf Bonn} the
correct semantic unit {\tt goal\-city:Bonn} was replaced by {\tt
goal\-city:Ber\-lin} in the parser output, thus being $SU_S=1$.  This
yields a concept accuracy of 50\% by calculating $CA =
100\left(1-\frac{1}{2}\right) \% = 50\%$.

The example shows that beside its ability to judge the parser
performance on a semantic level, CA is also an adequate measure for
evaluating robust parsers which allow partial analysis. This is a
distinguishing feature of CA in comparison with binary measures like
sentence recognition accuracy. In such approaches a system output
either totally agrees with a reference answer or it is counted as a
total failure. Concept accuracy on the other hand is able to measure
the {\em degree} of system understanding. In the above example, 50\%
CA expresses the fact that the chain comprising word recognizer and
parser was able to extract half of the propositional content from the
input utterance.

\subsection{Word Accuracy vs. Concept Accuracy}
\label{WA&SA}

The example shown in the previous section illustrates that the
relationship between WA and CA cannot be predicted
systematically. Both measures can differ considerably because WA does
not make a difference between {\em filler} words and semantically
relevant words. For example, WA in~(\ref{ExSA}) is 75\% (only 1
substitution error), whereas CA is only 50\%. This is explained by the
fact that the substituted city name forms the semantic core of the
{\tt goalcity}-concept which is misunderstood as a whole in
consequence. The opposite case is illustrated by example~(\ref{WASA})
where $WA=66.7\%$ but $CA=100\%$ because the misrecognitions did not
concern the parts relevant for understanding.
\begin{example}
\label{WASA}
\begin{tabular}{|l|ll|}\hline
Spoken: & {\sf I want to go} & {\sf to Berlin}\\
REF:    &                    & {\tt goalcity:Berlin}\\
\hline
Recog.: & {\sf I wonder go} & {\sf to Berlin} \\ 
HYP:    &                   & {\tt goalcity:Berlin}\\
\hline
\end{tabular}
\end{example}
The example shows that it is possible to achieve perfect utterance
understanding with less than perfect word recognition.  This happens
when misrecognitions only affect semantically irrelevant (in our
domain) filler words.  On the other hand, if recognition errors occur
within parts that are relevant for understanding an utterance, CA may
become lower than WA. This relationship between WA and CA was
investigated in the experiments we describe in section~\ref{RESULTS}.
These experiments were performed with the evaluation environment and
the data described in the next section.

\section{EVALUATION ENVIRONMENT}
\label{EVALENV}

\begin{figure}[tb]\centering
\mbox{\psfig{figure=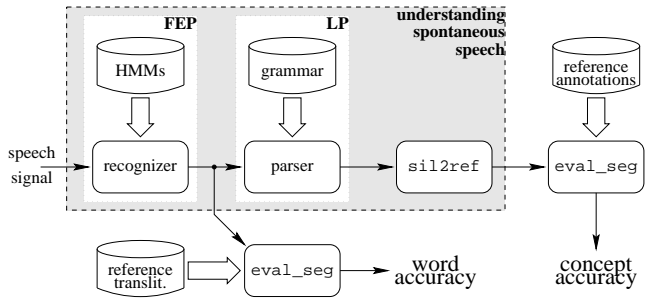,width=\columnwidth}}
\caption{Architecture of the automatic evaluation system.}
\label{EvalEnv}
\end{figure}

We implemented a test environment which can automatically calculate
the concept accuracy of the parsing results for a given semantically
annotated test corpus.  The architecture of our automatic evaluation
system is outlined in Figure~\ref{EvalEnv}.

The test corpus consists of a set of test cases, which are either
transliterations of the spoken utterance or word recognition results.
In the first case the environment is used for evaluating the
linguistic component alone, in the latter case word recognizer (FEP)
and linguistic processing component (LP) are evaluated together.  In
both cases each test sentence is combined with a semantic reference
annotation in the form of attribute-value pairs shown above.  The test
cases are handed over sequentially to the parser which tries to
analyze it with respect to its knowledge base, i.e. the grammar.  At
the moment we use a robust chart parser~\cite{Meckl:95} which selects
a set of partial results from the chart if no complete analysis can be
found.  This parser uses a highly lexicalized unification grammar
based on the UCG formalism~\cite{Zeevat:88}.  The strict modularity of
the evaluation environment allows an easy replacement of test data as
well as of the linguistic processing component. Thus, although we use
the evaluation programme mainly for progress evaluation, it can also
be used for comparative evaluation of alternative implementations of
the lingusitic component.  The only requirement is that the components
generate comparable results in the semantic interface language ({\sc
Sil},~\cite{Sil:90}) used in our dialogue system.  In order to compare
these complex parsing result structures with the much simpler
reference annotations, we implemented a (domain specific) module {\tt
sil2ref} which maps between {\sc Sil} and the annotated semantic
units.  Finally, the parsing results and the semantic annotations are
compared by calculating the Levenshtein distance by programme {\tt
eval\_seg}.  The resulting concept accuracy is reported
(cf. Figure~\ref{EvalEnv}).

\section{EXPERIMENTS \& RESULTS}
\label{RESULTS}

In our evaluation experiments we wanted to examine the relations
between WA and CA, in order to see if improvement of the word
recognizer (and thus WA) also leads to improvement of concept accuracy.
Therefore several evaluation tests were run.  Based on the same speech
material we run the recognizer with different parameter settings,
resulting in differences in word accuracy (and processing speed).
These word chains have been processed by the linguistic processor and
corresponding figures for WA and CA were calculated.

Evaluation was performed on a test corpus collected while the system
was accessible via the public telephone network~\cite{EckertNoeth:95}.
1092 dialogues with (naive) users were recorded.  We recorded the word
recognizer output, the transliterations and the semantic annotation for
each utterance were done manually.
  Table~\ref{corpus} gives
an overview of the test corpus.

\begin{table}[h]\centering
\begin{tabular}{|l|r|} \hline
Total number of dialogues &  1092 \\ \hline
Total number of utterances & 10114 \\ \hline
Total number of words & 33477 \\ \hline
Total number of semantic units & 14584 \\ \hline
Different classes of semantic units & 38 \\ \hline
\end{tabular}
\caption{Figures of the test corpus.}
\label{corpus}
\end{table}

The first step was to evaluate the linguistic component of the system
on its own, in order to measure the (semantic) coverage of the
grammar.  The resulting figure for CA reflects the grammars ability to
extract the meaning of an utterance and thus its adequacy for the
given domain.  For this purpose, CA was computed using the
transliterations as input to the parser and comparing the resulting
semantic representation with the reference annotation.  We achieved a
linguistic coverage of 92.8\% for spontaneous speech.

In order to examine the influence of different recognizer parameters
on the systems concept accuracy, several experiments were carried out.
The recognizer parameter to be altered was the beam width.  For each
parameter setting the recognizer was run on the recorded 10114
utterances of the corpus.  Concept accuracy was then measured using
the resulting recognizer output as input to the parser.
Table~\ref{evalres} shows the resulting marks for WA and corresponding
CA.

\begin{table}[h]\centering
\begin{tabular}{|l|r|r|r|r|r|r|}
\hline
 WA	& 48.8 & 65.7 & 72.9 & 77.5 & 83.0 & 84.9\\
\hline
 CA	& 46.7 & 61.9 & 68.2 & 73.0 & 78.5 & 79.8\\
\hline
\end{tabular}
\caption{Resulting marks for WA and CA when altering the
recognizer beam width.}
\label{evalres}
\end{table}

\begin{figure}[tb]\centering
\mbox{\psfig{figure=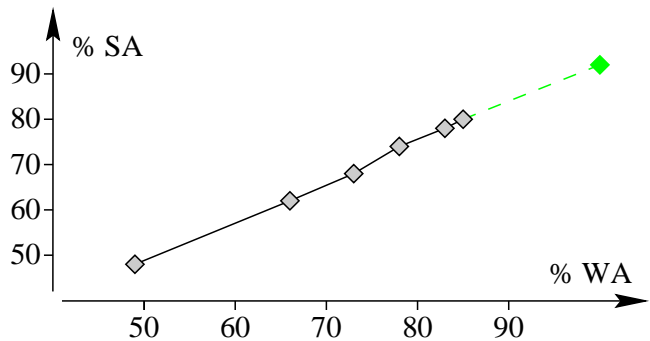,width=\columnwidth}}
\caption{Relationship between word accuracy and concept accuracy.}
\label{graphics}
\end{figure}

Table~\ref{evalres} shows that the marks for WA and CA correspond
closely.  This means that in our case the misrecognition in the
acoustic front end processor affects content words and filler words by
the same amount.  Moreover, we can see that the linguistic processor
does not suffer from misrecognition of a few words.  The parser has to
be judged as extremely robust against recognition errors as well as
phenomena of spontaneous speech.  Figure~\ref{graphics} shows the
nearly linear relation between word accuracy and corresponding
concept accuracy.

In our case we can make the assumption that word accuracy is a
suitable indicator for concept accuracy in a spoken dialogue system:
recognizer and parser are well matched for their tasks and cooperate
smoothly.

\section{SUMMARY}

In this paper we have shown an approach for the automated evaluation
of an understanding module for spontaneous speech.  This module
consists of an acoustic recognizer and a linguistic processor.  The
resulting semantic content of each utterance is compared automatically
with reference annotations, mimicking the evaluation of a word
recognizer alone.  Accordingly, the measure for a speech understanding
system is called {\em concept accuracy}.

With our evaluation setup we are able to document improvements in one
of our modules in an automated way.  Thus, we are not only able to
optimize isolated modules, but the whole understanding system.
Experiments show that our parser is robust in the sense that we
observe a nearly linear relation between WA and CA.

Further work will be commited to adjust parser parameters.  Eventually
we hope to increase CA beyond WA.

\section{ACKNOWLEDGEMENTS}
Part of this work was carried out in the project SYSLID which is
funded by the Daimler-Benz research institute in Ulm.  Part of this
work was supported by the German Research Foundation (DFG) under
contract number 810 830-0.

\nocite{Hildebrand95:SUB}

\bibliographystyle{plain}

\end{document}